# Breast Cancer Detection Using Multilevel Thresholding

Y.Ireaneus Anna Rejani[+],
[+] *Noorul Islam College of Engineering,Kumaracoil,,*
*Tamilnadu, India.*

Dr.S.Thamarai Selvi[*]
*Professor&Head, Department of Information and*
*technology, MIT, Chennai, Tamilnadu,* India.

*Abstract*— **This paper presents an algorithm which aims to assist the radiologist in identifying breast cancer at its earlier stages. It combines several image processing techniques like image negative, thresholding and segmentation techniques for detection of tumor in mammograms. The algorithm is verified by using mammograms from Mammographic Image Analysis Society. The results obtained by applying these techniques are described.**

## 1. INTRODUCTION

Breast cancer is one of the leading causes of cancer related death among women. The death rate can be reduced if the cancer is detected at its early stages. Early diagnosis and treatment increases the chance of survival. Early breast cancer detection requires periodical, readings of mammograms. Women over 40 years of age and those who have family history are recommended to take mammograms regularly for screening. At present, mammogram readings are performed by radiologists and mammographers, who visually examine mammograms for the presence of deformities that can be interpreted as cancerous changes. Manual readings may result misdiagnosis due to human errors caused by visual fatigue. To improve the diagnostic accuracy and efficiency of screening mammography computer aided diagnosis techniques are introduced.

The main aim of this work is the detection of cancer from mammograms. The mammograms suspicious for cancer are found out for more detailed examination by the attending physicians. There are several image processing methods proposed for the detection of tumors in mammograms. Although there are various tumor detection algorithms in the literature, the detection rate is still not high. Our algorithm is implemented using the concept of thresholding, segmentation and then finally checking the roughness value to identify tumor.

Image segmentation is typically used to locate objects and boundaries in images. After segmentation we get the required portion of the image. The segmented output may or may not be a tumor. The segmented output may be a fatty tissue. To confirm this, calculation of roughness value (D) is needed. Roughness of the image will be varying pixel to pixel. For tumor affected region the roughness value lies between 2 to 3.For other regions the roughness value will be less than 2 or it will be greater than 3.By this way the segmented output is confirmed whether it is a Tumor or not.

Digital mammography is a technique for recording x-ray images in computer code instead of on x-ray film, as with conventional mammography. The images are displayed on a computer monitor and can be enhanced (lightened or darkened) before they are printed on film. Images can also be manipulated; the radiologist can magnify or zoom in on an area. This screening will generate large number of mammograms to be determined by a small number of radiologists resulting in misdiagnosis due to human errors caused by visual fatigue. The sensitivity of human eye decreases with increasing number of images. Hence, it may be helpful for a radiologist, if a computer-aided system is used for detection of tumors in mammograms. Computer-aided detection (CAD) involves the use of computers to bring suspicious areas on a mammogram to the radiologist's attention. It is used after the radiologist has done the initial review of the mammogram.

There are several image processing methods proposed for detection of tumors in mammograms. In some cases the primary objective was to enhance the mammograms[1];in other cases,[2];[3];[4];[5];[6]researchers have concentrated on identifying areas in mammograms that may contain cancerous changes. Steps have been taken [8];[9];[10],to fully automate mammogram analysis. Various technologies such as wavelet based image denoising[11];multiresolution based image processing[13]and Markov random field(MRF)[14],have been used Even though many algorithms are available for tumor detection the detection rate is still not high. This is due to the high variance in size and shape of the tumors, and also due to the disturbance (noise) from the fatty tissues, veins and glands.

## 2. THEORY

### A. Image Negative

The negative of an image with gray levels in the range [0,L-1] is obtained by using the negative transformation shown in the figure1.It is given by the expression

$$s=L-1-r \qquad (1)$$

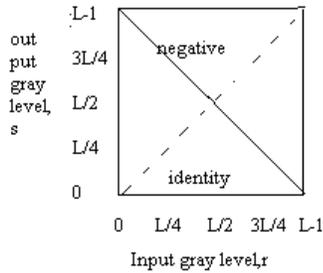

Figure 1. Image Negative

The digitized mammogram is applied for negative transformation. The complement of the image is obtained in this stage which is used for further processing. In image negative, the gray value of the original mammogram is complemented using the following mapping function

$$O(x,y) = 255 - I(x,y) \qquad (2)$$

### B. Threshold Value

After inverting the image, thresholding is done. The main goal of the thresholding is to classify each image pixels into two categories (eg.Foreground and Background). Thresholding is the process of picking up a fixed gray scale value and then to classify each image pixel by checking whether it lies above or below this threshold value.

Setting a threshold value for an image can be done by trial and error method. Threshold value is not same for images; it will be varying from one image to another. Hence the threshold can be called as Adaptive Threshold.

Thresholding is done to extract the portions of the image whose pixel intensity value is greater than the fixed threshold value of that particular image. Then the resultant image is subjected to the segmentation process.

### C. Segmentation

Image segmentation is the partitioning of an image into several constituent components. Segmentation should be stopped when the objects of interest in an application have been isolated. Segmentation distinguishes objects from background.

### D. Calculation of the Roughness value (D):

Fractals are rough or fragmented geometric shape that can be subdivided into parts, each of which is approximately a reduced copy of the whole. In fractal analysis, the fractal dimension measures the roughness of a block. Generally, an image is subdivided into N x N blocks and the fractal dimension is calculated for each block. Since in this work, a mammogram is used for cancer detection, the roughness of each and every pixel has to be taken. Use of fractal analysis reduces the search region. The area of the fractal surface canbe expressed as:

$$A_r = k \times r^{2-D} \qquad (3)$$

Where

    $A_r$ - Surface Area
    r   - Ruled Area
    k   - Scaling constant
    D  - Roughness of the region

Blanket method is used to calculate D.

$$Log\,(A_r) = (2\text{-}D)\,log(r) + k' \qquad (4)$$

For a surface D is between 2-3. The larger the D is, rougher the surface. For all subdivided blocks of a mammogram, the blocks that have smooth surface or a very rough surface are discarded.

### E. Parameters for Tumor Detection:

The features selected in our approach to locate the regions that are suspicious of tumors are given as follows.

*Area A:* This parameters is the total number of pixels with in a certain extracted region.

*Compactness cmp:* This quantity reflects the shape of the given region and equal to

Cmp = (Area of the given region)/ (Area of the smallest rectangle circumscribe the given region)

*Mean gradient with in current region – Mwg:* This parameter measures the average gradient of each pixel in the given region.

$$Mwg = (1/N) \; {}^{N}_{k=1} \; (g_k) \qquad (5)$$

Where N equals the total number of pixels with in the given region and $g_k$ is the gradient at each pixel k.

*Mean gradient of region boundary – mg:* This parameter indicates the sharpness of the region boundary.

$$Mg = (1/N') \; {}^{N}_{k=1} \; (g_k') \qquad (6)$$

Where N' equals the total number of pixels on the boundary of the given region and $g_k'$ is the gradient along the boundary of the given region.

*Gray value variance – var:* var measures the smoothness of the given region

$$Var = ((1/N) \quad (i, j) \in A\,(X\,(i, j) - X')^2)^{1/2} \quad (7)$$

$$X' = (1/N) \quad (i, j) \in A\,X\,(i, j) \qquad (8)$$

Where X (i, j) is the gray level of each pixel with on region, A and N is the total number of pixels in the region.

*Edge Distance Variance* - edv: edv measures the shapes of the shape of the region and its rotational symmetry

$$\text{Edv} = ((1/N) \quad {}^{N}_{k=1}(dk - d')^2)/d' \qquad (9)$$

Where $d_k$ represents the distance from pixel k on the edge to the center of the    region and d' is the mean value of all edge distances.

*Mean Intensity Difference* – diff: This parameter measures the gray value difference between the value inside the region and those outside the region but inside the smallest rectangle cover of the region.

$$\text{Diff} = (1/N_a) \quad (i, j) \in A \ X\ (i, j) - (1/N_c) \quad (i, \ j) \in C$$
$$X(i, j) \qquad (10)$$

Where $N_a$ is the total number of pixels in region A, $N_c$ is the total number of pixels in a rectangle region C, which represents the region pixels covered by the        rectangle but not inside the region A.

### 3. DEVELOPMENT OF THE ALGORITHM

#### F.  Preprocessing steps

The techniques done in the preprocessing steps were the inverting and thresholding of the mammogram image. The input mammogram is shown in figure as given below

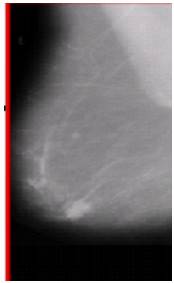

Figure 2  Input image

#### G.  Inverting steps

Intensity of the first pixel is taken. Subtract the maximum intensity value (255) from the intensity value taken from the pixel. Then repeat the first two steps for all the pixels in the image to invert the image.

After inverting all the pixels value, we get inverted image as shown in figure

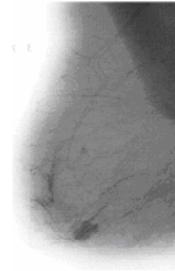

Figure 3 Inverted image

#### H.  Setting the threshold value

From the inverted image, we have to find the threshold value. Normally the threshold value can be found by trial and error method or from the histogram of the image. After finding the threshold value, we have to apply for the image. Threshold is used to classify each image pixel by checking whether it lies above or below this threshold value. Then finally taking the object from the image as shown in the figure

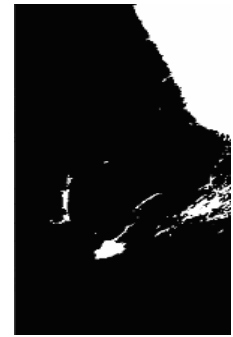

Figure 4 Thresholded image

#### I.  Steps for segmentation

Segmentation refers to the process of partitioning a digital image into multiple regions (sets of pixels).segmentation can be done by the following steps.

Now the pixels having same intensity value are grouped into regions. Image was taken into consideration and checked whether they had dissimilarities within themselves. If there were dissimilarities those regions were again subdivided. After each split, the adjacent regions were compared and merged if there exists any similarity between them. This process was continued until there was no further splitting or merging possible. Finally the roughness value of each region is calculated. If the roughness value lies between 2 to 3, then that region is finally segmented as shown in the figure given below

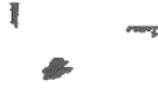

Figure 5 Final segmented image

### J.  Tumor classification

From the segmented output, the area of the segmented image is calculated. Similarly compactness, variance etc. are calculated from the segmented image. These features are used for the classification of the segmented area into tumor or normal.

### K.  Results

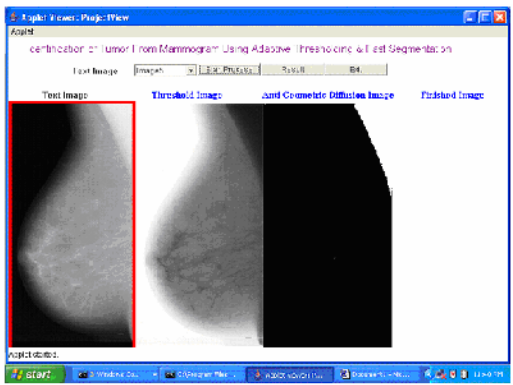

Figure 6 For Normal image

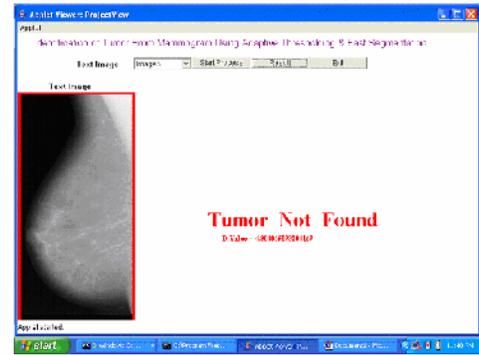

Figure 7 Result for normal image

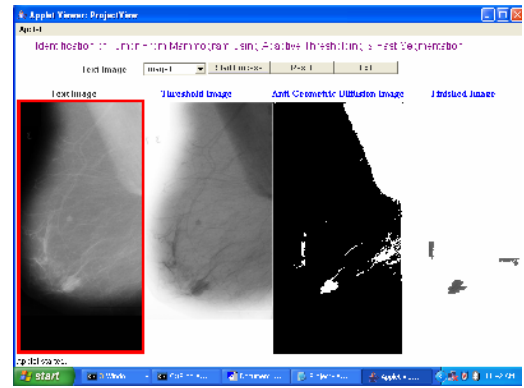

Figure 8 For affected image

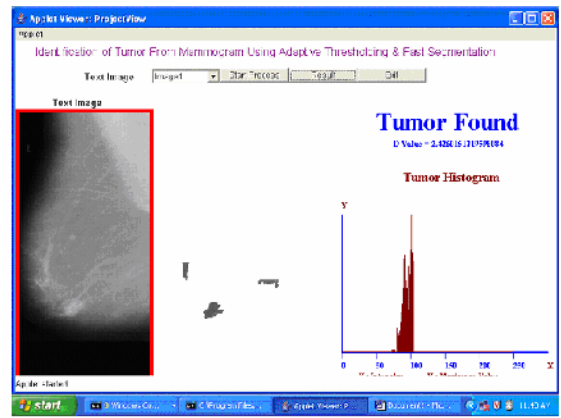

Figure 9 Result for affected image

### 4.CONCLUSION

This algorithm is verified using a database from MIAS. The fractal dimension used to process the mammograms was between 2.4 and 2.75.  A 3 level DWT decomposed image had been chosen .Three level decomposition reduces an image of size 1024×1024 to 128×128. The results of this algorithm are the identification of tumor from normal ones.

In this project, we have described an algorithm that acts as a preprocessor for marking out the suspicious tumor regions in the mammogram to increase the segmentation accuracy, fractal analysis and Adaptive Thresholding is used for the segmentation initialization procedure. Fast Segmentation is used for the Final segmentation. During the Classification the properties of the tumor are calculated. The result shows that Adaptive Thresholding and Fast Segmentation Algorithm is efficient and successful. This algorithm acts as an assistant to radiologists in detecting tumors in mammograms.

### ACKNOWLEDGEMENTS

The authors are thankful to The International Cancer Research Institute ,Neyoor, Tamilnadu ,India.

### AUTHORS PROFILE

1.Y.Ireaneus Anna Rejani is working as assistant professor in Noorul Islam College of Engineering,Kumaracoil,, Tamilnadu, India.Her area of research is Neural networks.

2. Dr.S.Thamarai Selvi is working as Professor&Head, Department of Information and technology, MIT, Chennai, Tamilnadu, India.She is having vast teaching and research experience.Her area of interest is neural network and grid computing.